\documentclass[letterpaper, 10 pt, conference]{ieeeconf}
\IEEEoverridecommandlockouts
\usepackage{graphics} 
\usepackage{wrapfig} 
\usepackage[linesnumbered,ruled]{algorithm2e}
\graphicspath{{img/}}
\usepackage{epsfig} 
\usepackage{amsmath} 
\usepackage{amssymb}  
\usepackage{amsthm}
\usepackage{bigints}
\usepackage{pifont}
\def\-{\raisebox{.75pt}{-}}
\usepackage{mathtools}
\usepackage{cite}
\usepackage[makeroom]{cancel}

\newcommand\copyrighttext{%
	\footnotesize This work has been submitted to the IEEE for possible publication. Copyright may be transferred without notice, after which this version may no longer be accessible.}
\newcommand\copyrightnotice{%
	\begin{tikzpicture}[remember picture,overlay]
	\node[anchor=south,yshift=10pt] at (current page.south) {\fbox{\parbox{\dimexpr\textwidth-\fboxsep-\fboxrule\relax}{\copyrighttext}}};
	\end{tikzpicture}%
}

\usepackage[table]{xcolor} 

\usepackage[center]{subfigure}
\usepackage{multirow}
\usepackage{hyperref}
\usepackage{booktabs}
\usepackage{subfigure}
\usepackage{epstopdf}
\usepackage{textcomp}
\usepackage{makecell}
\usepackage{bm}
\usepackage{caption}
\usepackage{algpseudocode,algorithm} 
\usepackage{color, soul}
\usepackage{mathptmx}
\usepackage[11pt]{moresize}
\usepackage{hhline}
\usepackage{tikz}
\usepackage{textcomp}
\def\BibTeX{{\rm B\kern-.05em{\sc i\kern-.025em b}\kern-.08em
    T\kern-.1667em\lower.7ex\hbox{E}\kern-.125emX}}

\newcommand{\subparagraph}{}

\title{\huge Leveraging Uncertainties for Deep Multi-modal Object Detection in Autonomous Driving
\thanks{$^1$ Robert Bosch GmbH, Corporate Research, Driver Assistance Systems and Automated Driving, 71272 Renningen, Germany.}
\thanks{$^2$ Institute of Measurement, Control and Microtechnology, Ulm University, 89081 Ulm, Germany.}
\thanks{$^3$ Department of Mechanical Engineering, Karlsruher Institute of Technology, 76049 Karlsruhe, Germany.}
\thanks{$^*$ Work done while at Bosch.}
}

\author{Di Feng$^{1,2}$, Yifan Cao$^{3*}$, Lars Rosenbaum$^1$, Fabian Timm$^1$, Klaus Dietmayer$^2$}

\let\oldtwocolumn\twocolumn
\renewcommand\twocolumn[1][]{%
	\oldtwocolumn[{#1}{
		\begin{center}
			\includegraphics[width=0.96\textwidth]{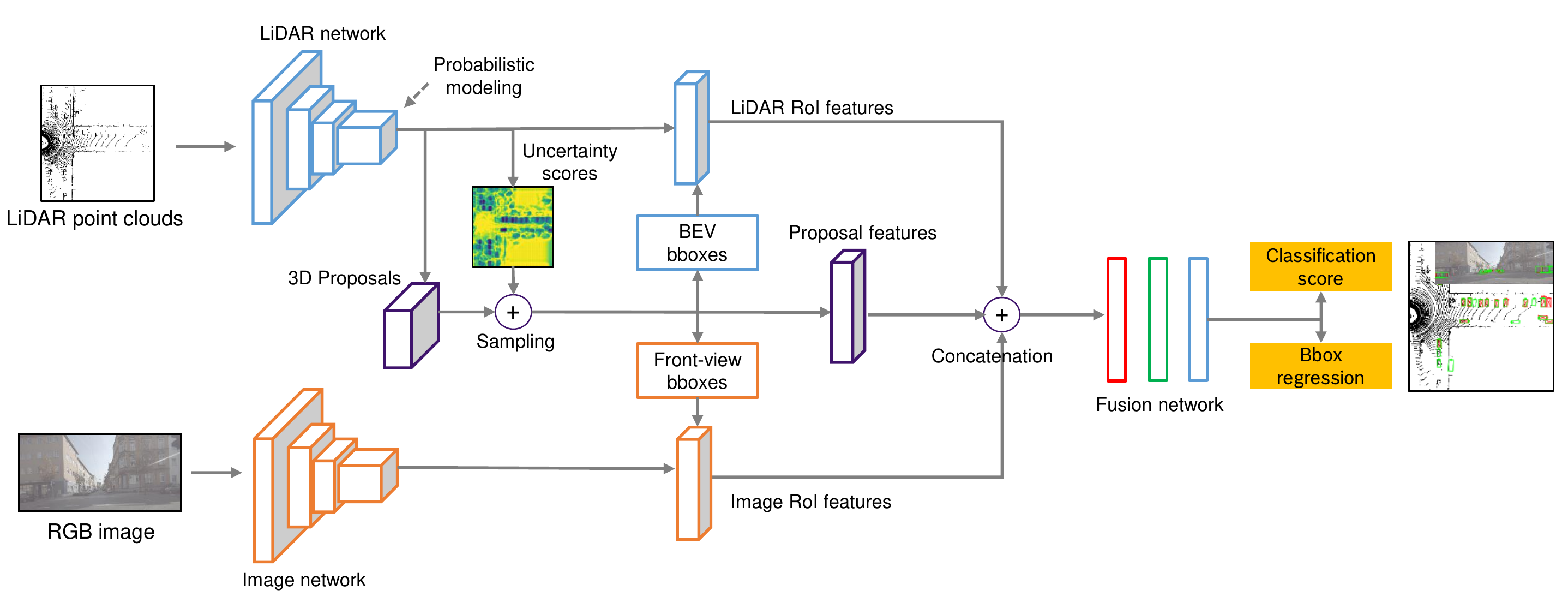}
			\captionof{figure}{Our proposed two-stage multi-modal object detection network which fuses LiDAR point clouds and RGB camera images. First, LiDAR and camera inputs are preprocessed by two sensor-specific networks respectively. Then, the 3D proposals predicted by the LiDAR network are used to extract regional LiDAR features and camera features. Finally, the regional features are combined by a light-weight fusion network for the final 3D bounding box regression and object classification. We do probablistic modeling in the LiDAR network, and introduce an uncertainty sampling mechanism during training to enhance the fusion network robustness.}\label{fig:framework}
		\end{center}
	}]
}

\begin{document}
\maketitle
\begin{abstract}
This work presents a probabilistic deep neural network that combines LiDAR point clouds and RGB camera images for robust, accurate 3D object detection. We explicitly model uncertainties in the classification and regression tasks, and leverage uncertainties to train the fusion network via a sampling mechanism. We validate our method on three datasets with challenging real-world driving scenarios. Experimental results show that the predicted uncertainties reflect complex environmental uncertainty like  difficulties of a human expert to label objects. The results also show that our method consistently improves the Average Precision by up to $7\%$ compared to the baseline method. When sensors are temporally misaligned, the sampling method improves the Average Precision by up to $20\%$, showing its high robustness against noisy sensor inputs.
\end{abstract}

\copyrightnotice


\section{Introduction}\label{sec:introduction}
A driverless car is usually equipped with multiple onboard sensors, such as video, LiDAR, and Radar sensors, in order to build a robust and accurate scene understanding. An object detection system that can exploit the complementary properties of different sensing modalities is crucial for safe autonomous driving.
 
As a powerful tool for learning hierarchical feature representations and complex transformations, deep learning has been widely applied to computer vision tasks. In this regard, many methods have been proposed recently which employ deep learning to fuse multiple sensors for object detection in autonomous driving~\cite{feng2019deep}. Typical methods such as MV3D~\cite{chen2017multi} and AVOD~\cite{ku2017joint} have achieved promising results on the standard open datasets, e.g. KITTI~\cite{Geiger2012CVPR}, with perfectly-aligned sensors and good weather conditions. However, those methods are not robust against temporal and spatial sensor misalignment that might occur during driving. Even a small spatial sensor displacement of $0.2$m has been shown to drastically degrade the network performance~\cite{shin2019roarnet}. It is still a challenge to improve the network's robustness against noisy sensor data.

Reliable uncertainty estimation in object detection networks provides extra information to support predictions, and has the potential to improve other modules such as motion planning~\cite{banzhaf2018footprints}. Previous work has studied uncertainties in object detectors that only employ a single sensing modality such as LiDAR point clouds~\cite{feng2018towards, feng2018leveraging, feng2019deep2, feng2019can,meyer2019lasernet} or RGB camera images~\cite{miller2017dropout,le2018uncertainty,harakeh2019bayesod,he2019bounding}. To the best of our knowledge, uncertainty estimation has not been introduced to multi-modal object detection yet. Furthermore, though previous studies have illustrated that uncertainties reflect environmental noises~\cite{feng2018leveraging} or data distribution~\cite{miller2017dropout}, there has been no study on how they reflect more complex attributes, which can be measured by the difficulties of a human expert to correctly label an object. Ideally, a probabilistic detector should assign high uncertainties to objects which human oracles also find difficult to label.

In this work, we propose a probabilistic two-stage multi-modal object detection network that fuses LiDARs and RGB cameras. We explicitly model classification and regression uncertainties in the network, and study how they reflect labeling difficulties of a human expert. To do this, we build a dataset with labels for environmental effects (e.g. occlusion, number of points) as well as an estimate of the labeling difficulty. Regarding the labeling difficulty, human annotators assign objects to be either ``Unsure" or ``Sure". Statistical analyses show that our network produces higher uncertainties when detecting ``Unsure" objects compared to ``Sure" objects after correction for other environmental effects. Afterwards, we leverage those uncertainties to improve the detection accuracy and the network's robustness, especially in the sensor misalignment situation. The method works by optimizing the fusion network with the training data sampled by the estimated probability distributions. We evaluate our method on multiple datasets with real-world driving scenarios, including two open datasets (KITTI~\cite{Geiger2012CVPR} and NuScenes~\cite{nuscenes2019}) and our self-recorded Bosch dataset. 

The \textbf{Contributions} in this paper are three-fold:
\begin{itemize}
	\item We explicitly model classification and regression uncertainties in a multi-modal object detection network, and study how uncertainties are affected by complex environmental uncertainties like labeling difficulty.
	\item We leverage uncertainties to improve the detection accuracy and the network's robustness, especially when sensors are temporally misaligned.
	\item We validate our proposed method on three real-world datasets.  
\end{itemize}

\section{Related Work}
\label{sec:related_works}
In this section, we briefly summarize deep learning methods for multi-modal object detection and uncertainty estimation. For a more comprehensive overview, we refer interested readers to the survey paper~\cite{feng2019deep}.

\subsection{Multi-modal Object Detection}
When combining multiple sensors for object detection, RGB cameras and LiDARs are the most common sensors reported in the literature~\cite{chen2017multi,qi2017frustum,ku2017joint,liang2018deep}. Some works propose to combine RGB images with thermal images~\cite{wagner2016multispectral}, LiDAR point clouds with HD maps~\cite{liang2019multi}, as well as RGB images with Radar points~\cite{chadwick2019distant}.
Modern multi-modal object detection networks follow either the two-stage or the one-stage pipeline. The variety of network architectures provide many options for sensor fusion. For instance, in the two-stage pipeline, different sensors can be combined at the first stage~\cite{ku2017joint,liang2019multi}, or at the second stage after regional proposal generation~\cite{chen2017multi,qi2017frustum,shin2019roarnet}. In the one-stage pipeline, sensing modalities can be fused at one specific layer~\cite{chadwick2019distant} or multiple layers~\cite{liang2018deep,wang2017fusing}. Typical fusion operations include feature concatenation, element-wise average mean, and ensembling. As discussed in~\cite{feng2019deep}, we do not find conclusive evidence that one fusion scheme is better than the others, and the fusion performance is highly dependent on the network architectures and datasets.

In this work, we propose a two-stage object detection network that combines RGB cameras and LiDARs. We first extract sensor-specific features using two different backbone networks, and then perform feature concatenation after the regional proposal generation (Fig.~\ref{fig:framework}). 

\subsection{Uncertainty Estimation and Probablistic Object Detectors}
There are many ways of estimating predictive probabilities in supervised deep networks. Bayesian Neural Networks (BNN)~\cite{mackay1992practical} place priors over the network weights, and infer their posterior distributions through variational inference~\cite{blundell2015weight,graves2011practical} or Monte Carlo sampling~\cite{Gal2016Uncertainty}. Deep Ensembles~\cite{lakshminarayanan2017simple} obtains predictive probabilities from an ensemble of networks with the same architecture but different training initializations. Uncertainty propagation methods approximate the variance in each activation layer and propagate uncertainties through networks~\cite{gast2018lightweight,postels2019sampling}. Direct-modelling approaches assume certain distributions over the network outputs. Networks are then trained to directly predict output distributions~\cite{kendall2017uncertainties,gast2018lightweight,choi2018uncertainty} or their higher-order conjugate priors, such as estimating the Dirichlet prior for the multinomial distribution in the classification task~\cite{sensoy2018evidential}, or the Gaussian prior on the mean and an Inverse-Gamma prior on the variance for the Gaussian distribution in the regression task~\cite{amini2019deep}.

We can capture two types of uncertainties in an object detection network: the epistemic uncertainty and the aleatoric uncertainty~\cite{feng2018towards}. The former reflects the model's capability for describing data, and can be explained away given enough training data; the later captures observation noises inherent in environments or sensors~\cite{kendall2017uncertainties}. Previous studies have leveraged epistemic uncertainty to improve detections in open-set conditions~\cite{miller2017dropout} or boost training efficiency in the active learning setting~\cite{feng2019deep2}. Other works have shown that modelling aleatoric uncertainty, especially in the bounding box regression task, can greatly improve the detection accuracy~\cite{feng2018leveraging,he2019bounding,meyer2019lasernet,meyer2019learning} and reduce False Positives~\cite{le2018uncertainty,Choi_2019_ICCV}. Miller \textit{et al.}~\cite{miller2018evaluating} and Harakeh \textit{et al.}~\cite{harakeh2019bayesod} have found that the merging strategy, such as Non Maximum Suppression (NMS), significantly influence the uncertainty estimation. Feng \textit{et al.}~\cite{feng2019can} identify uncertainty miscalibration problems in a one-stage object detection network. They follow\mbox{~\cite{guo2017calibration}} and calibrate uncertainties to correctly estimate the prediction error within the training data distribution.

In this work, we use the direct-modelling approaches to explicitly model aleatoric uncertainties for both classification and regression tasks. We employ uncertainties to improve the detection accuracy and the network's robustness against the sensor temporal misalignment.

\section{Methodology}
\label{sec:methodology}

\subsection{Network Architecture}\label{subsec:architecture}
Fig.~\ref{fig:framework} illustrates our proposed two-stage multi-modal object detection network which fuses LiDAR point clouds and RGB camera images. Following~\cite{chen2017multi,ku2017joint}, the LiDAR point clouds are discretized and projected onto the Bird's Eye View (BEV) plane, because this representation has been shown to be very effective in 3D perception~\cite{feng2019deep}. The input signals are first processed separately by sensor-specific networks to extract high-level feature maps. The LiDAR network head also generates accurate 3D object proposals. Afterwards, these proposals are projected onto the BEV and front-view to extract Region of Interest (RoI) features from LiDAR and image feature maps respectively. Finally, RoI features are combined in a small fusion network for 3D bounding box regression and object classification. 

The network architecture is designed in a modular manner that eases adoption. In practice, we can directly leverage off-the-shelf pre-trained sensor-specific networks to process LiDAR and camera data (as we do in this work). We can also easily adopt the network with new sensors (e.g. Radar) by re-training only the light-weight fusion network, without affecting other modules. However, the fusion performance is limited by the LiDAR network. If a 3D proposal is not recognized, e.g. due to sparse LiDAR point clouds, the object within it can never be detected by the fusion network. A potential improvement is to generate 3D proposals from Radar or camera channels, which we leave as an interesting future work.

\subsubsection{Input and Output Encodings}\label{subsec:encodings}
Denote an input sample as $\mathbf{x}$, and a 3D proposal generated by the LiDAR network as $\mathbf{z}$. It includes the class label $c_z$ with the softmax score $s_z$, and the proposal's location $\mathbf{b}_z$, i.e. $\mathbf{z}=[c_z, \mathbf{b}_z]$. For brevity we only consider binary class ``Object" and ``Non-object", $c_z\in\{0,1\}$. We encode $\mathbf{b}_z\in \mathbb{R}^8$ as the center positional offsets on the horizontal plane ($dx$ and $dy$), proposal bottom positional offset $dz$, length, width, height at log scale ($\log(l)$, $\log(w)$, and $\log(h)$), as well as orientation ($\cos(\theta), \sin(\theta)$). The fusion network predicts $\mathbf{y}=[c_y, \mathbf{b}_y]$, where $c_y$ is the class label with the softmax score $s_y$, and $\mathbf{b}_y$ the 3D bounding box position. We encode $\mathbf{b}_y\in \mathbb{R}^8$ as the offsets to the region proposal prediction $\mathbf{b}_z$.

\subsubsection{Sensor-specific Networks and Fusion Networks}
We process LiDAR data and extract 3D proposals using PIXOR~\cite{yang2018pixor}, a state-of-the-art one-stage LiDAR object detector, with several modifications. We estimate the object's height instead of only predicting on the BEV plane, and explicitly model predictive probabilities, which will be introduced in Sec.~\ref{subsec:learning_with_probablity}. As for the RGB image data, we employ the Feature Pyramid Network~\cite{lin2017feature}, a well-performing image feature extractor. 
The fusion network combines LiDAR and image RoI features through concatenation, similar to~\cite{ku2017joint}. It also takes the proposal positions and softmax scores as inputs, because we find that these proposal features can improve the 3D bounding box regression. The fusion network consists of three fully connected layers, each with 256 hidden units and being followed by a dropout layer. 

\subsection{Learning with Probability}\label{subsec:learning_with_probablity}
Suppose we have pre-trained LiDAR and camera networks that produce 3D proposals $\mathbf{z}$ and RoI features for fusion. The standard approach to training the fusion network can be viewed from the maximum likelihood perspective, where we learn a set of network weights $\mathbf{w}$ that maximize the observation likelihood of the training data. We minimize the negative log likelihood by setting the loss function:
\begin{equation}
\mathcal{L}(\mathbf{w}) = -\log \big( p(\mathbf{y}|\mathbf{x},\mathbf{z}) \big).
\end{equation}
In the context of classification, $p(\mathbf{y}|\mathbf{x},\mathbf{z})$ is usually set to be the multinomial mass function, and $\mathcal{L}(\mathbf{w})$ is widely known as the cross-entropy loss. It can also be adapted to tackle the positive negative sample imbalance problem via the focal loss~\cite{lin2017focal}. As for the deterministic regression, we can assume $p(\mathbf{y}|\mathbf{x},\mathbf{z})$ as the Gaussian density function with fixed variance. The corresponding loss function is the $L_2$ loss. 

\begin{figure}[tpb]
	\centering
	\includegraphics[width=0.96\linewidth]{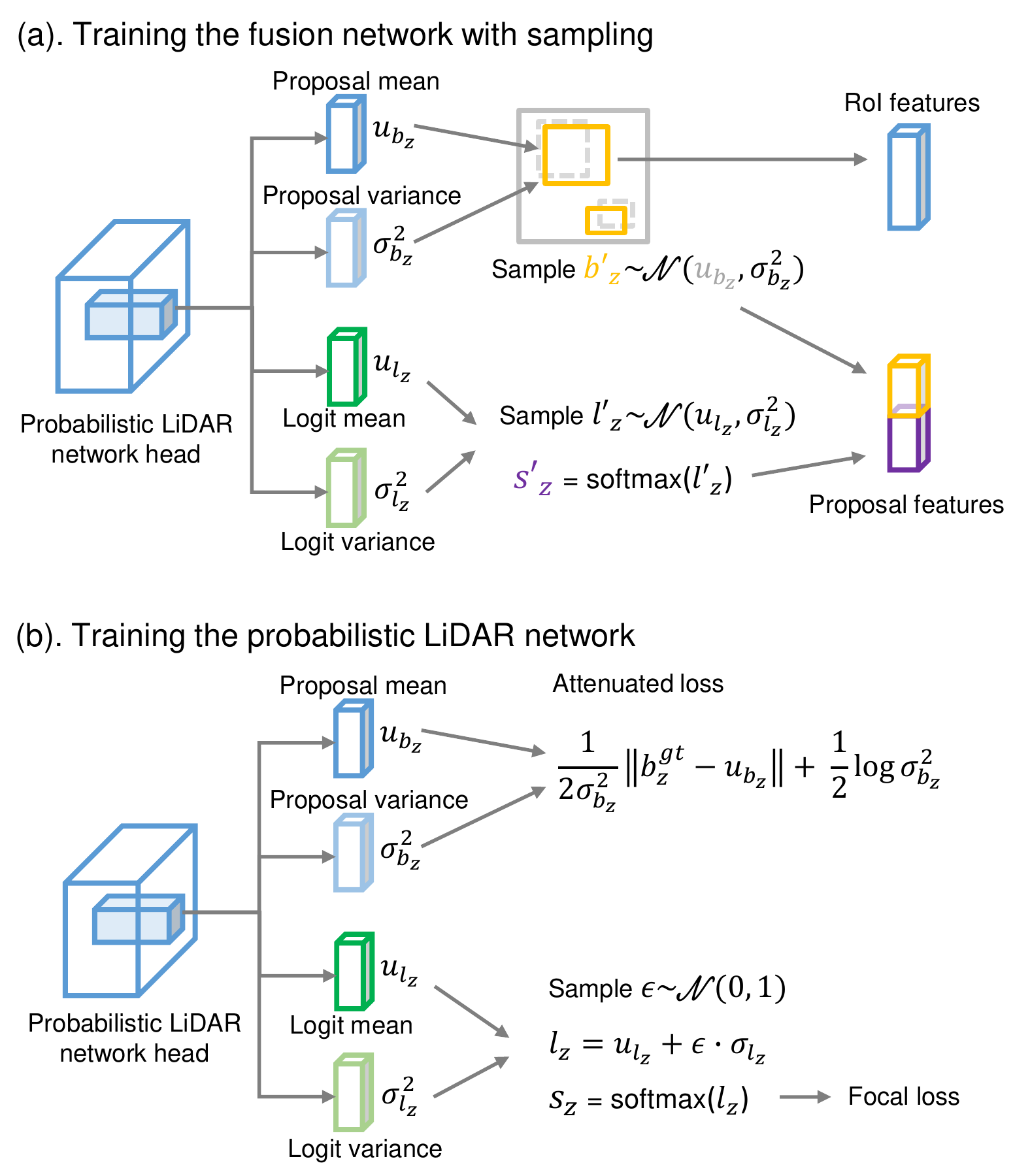}
	\caption{(a). Training the fusion network with sampling. We assume that the 3D proposals are Gaussian distributed, and propagate the sampled proposals based on predicted probability distribution to the fusion network; (b). Training the probabilistic LiDAR network. The network directly regresses the parameters of the probability distributions, which are incorporated in the loss function.}\label{fig:sampling}
\end{figure}

In this work, we incorporate the proposal distribution $p(\mathbf{z}|\mathbf{x})$ into the loss function:
\begin{equation}
\mathcal{L}(\mathbf{w}) = - \mathbb{E}_{ p(\mathbf{z}|\mathbf{x})}  [\log \big( p(\mathbf{y}|\mathbf{x},\mathbf{z}) \big)].
\end{equation}
Since an analytical solution is intractable, we approximate this loss function via sampling (as illustrated in Fig.~\ref{fig:sampling}(a)):

\begin{equation}
\text{Sample}\ \mathbf{z'} \sim p(\mathbf{z}|\mathbf{x}),\ \mathcal{L}(\mathbf{w})= -\log \big( p(\mathbf{y}|\mathbf{x},\mathbf{z'}) \big)
\end{equation}

This new training strategy brings two benefits. First, propagating proposal uncertainties to the fusion network helps to improve its robustness, as the network learns to handle proposals with small and big uncertainties. Second, sampling proposals can serve as a simple data augmentation method that aids generalization. 

In practice, we could pre-define a proposal distribution, such as the Gaussian distribution with fixed variance. In this work, we use predictive uncertainties from a \textit{pre-trained} probabilistic LiDAR network, which is more flexible because they can encode both the varying environmental noises and the network's inaccuracy for each proposal. We will illustrate how to model probability in the following section.

\noindent \textbf{Probabilistic Modelling:}
We explicitly model regression and classification uncertainties in our LiDAR network. For simplicity we will use the scalar notation instead of vector to introduce our method, e.g. $b_z$ is a regression variable in the vector $\mathbf{b}_z$. Fig.~\ref{fig:sampling}(b) shows the process. Following~\cite{feng2018leveraging}, we assume that each proposal regression variable is Gaussian distributed, i.e. $b_z \sim \mathcal{N}(u_{b_z}, \sigma^2_{b_z})$, with its mean $u_{b_z}$ being the network standard output, and its variance $\sigma^2_{b_z}$ (\textit{regression noise}) an auxiliary regression variable. We employ the attenuated loss proposed by~\cite{kendall2017uncertainties} to learn this probability distribution (see Fig.~\ref{fig:sampling}(b)). Similarly, we assume the distribution of softmax logit $l_z$ to be Gaussian, i.e. $l_z \sim \mathcal{N}(u_{l_z}, \sigma^2_{l_z})$, and add another output layer in the LiDAR network head to regress the logit variance $\sigma^2_{l_z}$ (\textit{classification noise}). Directly learning this variable is difficult~\cite{kendall2017uncertainties}. Instead, we sample a logit using the re-parametrization trick and transform it to the softmax score, which is used to calculate the final classification loss (see Fig.~\ref{fig:sampling}(b)).

It is noteworthy to mention that we train the probabilistic LiDAR network and the fusion network separately to favour the modular architecture design, as discussed in Sec.~\ref{subsec:architecture}. When optimizing the fusion network under the proposal uncertainties, we directly sample the proposal position $b_z'$, and indirectly propagate the softmax uncertainty by sampling the softmax logit $l_z'$ (see Fig.~\ref{fig:sampling}(a)). All sampling operations are not required during the inference. Therefore, this approach to directly modelling uncertainties brings almost no additional computational cost and parameters, as discussed in~\cite{feng2018leveraging}. In practice, we could also train the whole detector in an end-to-end fashion by employing the re-parametrization trick to the proposal variance $\sigma^2_{b_z}$. However, we do not find improvement on the detection accuracy using end-to-end training. 
\section{Experimental Results}
\label{sec:result}
The experimental results are structured as follows. In Sec.~\ref{subsec:uncerstanding_uncertainty}, we study what the predictive uncertainties used in the fusion network look like. We conduct statistical analyses and show that those explicitly-modelled uncertainties reflect complex environmental noises and the labelling uncertainty from human annotators. Afterwards, we show in Sec.~\ref{subsec:detection_performance} that the proposed uncertainty estimation and sampling mechanism improve the object detection performance across three datasets. Specifically, we observe that the proposed sampling mechanism is more robust than the fixed-sampling approach, because the predictive uncertainties encode useful information as shown in the first experiment. Finally, in Sec.~\ref{subsec:robustness_testing} we demonstrates the robustness of our method against sensor temporal misalignment problems.

\subsection{Setup}
\subsubsection{Datasets}
We validate our method on detecting objects of the ``Car" category in three real-world urban driving datasets recorded at different locations and with different sensor setups. 

\noindent \textbf{KITTI}~\cite{Geiger2012CVPR}: the dataset is recorded in Karlsruhe, a mid-sized city in Germany, only during daytime and on sunny days. Following~\cite{chen2017multi}, we split the training data of $7481$ frames into a \textit{train} set and a \textit{val} set, with approximately $50/50$ ratio. The network is optimized on the \textit{train} set and evaluated on the \textit{val} set.

\noindent \textbf{Bosch}: we also record data in several major cities in southern Germany with the vehicle setup similar to KITTI, but in much more diverse driving scenarios, such as night-drive and rainy or cloudy days. We follow KITTI and label the object truncation and occlusion using ordinal numbers. Besides, we ask annotators to label each object as either ``Unsure" or ``Sure". An object is ``Unsure", if the annotators find it difficult to define its ground truth label, such as box parameters. Such label enables us to study how the predictive uncertainties from our model reflect the labelling difficulties of a human expert. In the experiment, we randomly split the data into train drives ($8664$ frames) and test drives ($3028$ frames). 

\noindent \textbf{NuScenes}~\cite{nuscenes2019}: this large-scale dataset is recorded in Singapore and Boston, with rich complexity of traffic and weather conditions. Different from KITTI and Bosch datasets which use 64-lens LiDARs, NuScenes is equipped with 32-lens LiDARs~\cite{nuscenes2019}, making the LiDAR perception more challenging. Since the full dataset is quite large, we only use a small subset to do evaluation. We randomly select $100$ scenes in the full training data ($6022$ frames) to train the network, and the data in the NuScenes-teaser release for testing ($3962$ frames). 

\subsubsection{Implementation Details}
We assemble our multi-modal object detector following the modular design discussed in Sec.~\ref{subsec:architecture}. First, we leverage the pre-trained Feature Pyramid Network as the image backbone directly from Detectron~\cite{Detectron2018}, and pre-train the PIXOR-like LiDAR network with the SGD optimizer and the learning rate $0.02$, and set the step decay to be $0.75$ for every $30,000$ training steps. We train the LiDAR network with $140,000$ steps for KITTI, and $300,000$ steps for Bosch and NuScenes. Our LiDAR network achieves similar detection performance compared to the original PIXOR results~\cite{yang2018pixor}. Afterwards, we fix the sensor-specific networks, and train the fusion network with the ADAM optimizer and learning rate $10^{-5}$ ($120,000$ steps for KITTI, and $250,000$ steps for Bosch and NuSecenes). We find this strategy of long training with small learning rate makes the fusion network more stable. We proceed $1024$ 3D proposals with the highest classification scores to the fusion network during the training process, and reduce the number to $500$ during inference. Proposals out of the camera field of view are not considered for fusion. For KITTI and Bosch datasets, we use the LiDAR point cloud
within the range length $\times$ width $\times$ height = $[0,70]$m$\times[-
40,40]$m$\times[0,2.5]$m, and do discretization with $0.1$m resolution. For the NuScenes dataset we set the length up to $50$m due to LiDAR point cloud sparsity. All experiments are conducted using Titan X GPUs. The inference time reaches $5$fps.

\subsection{Understanding Uncertainties}\label{subsec:uncerstanding_uncertainty}
We study how uncertainties behave using the probabilistic proposals predicted by the LiDAR network on the Bosch test data. We measure the Shannon Entropy of softmax scores ($SE$), the total variance $\sigma^2_{b_z}$ of the regression noise, as well as the classification noise $U_{cls}$ by the positive logit variance $\sigma^2_{l_z}$. In the binary case, the shannon entropy is high for softmax scores close to 0.5 and low for scores close to 0 or 1.

\begin{figure*}[tpb]
	\centering
	\includegraphics[width=1\linewidth]{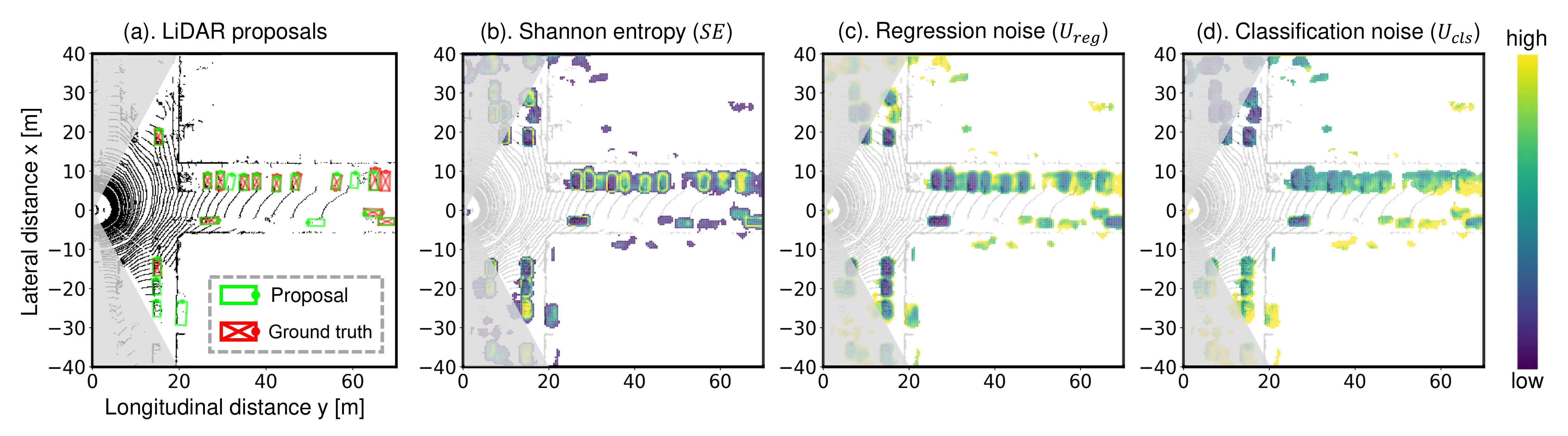}
	\caption{A detection example. We only visualize uncertainties (normalized to the same scale) at the regions with the softmax objectness score larger than $0.01$. Areas out of the camera field of view are shaded in grey.}\label{fig:uncertainty_visualization}
\end{figure*}

Fig.~\ref{fig:uncertainty_visualization} shows an example of how uncertainties are distributed in an input frame. We only visualize uncertainties at regions with positive softmax scores larger than $0.01$, because below this threshold regression uncertainties are dominated by random noises due to the training strategy (no regression loss is a region is not assigned to a ground truth). We observe that the proposals at the ``around-object" regions usually show higher Shannon Entropy scores than the ``in-object" regions (Fig.~\ref{fig:uncertainty_visualization} (b)), because those regions are on the boundary between objects and background. The regression and classification noises do not show this behaviour (Fig.~\ref{fig:uncertainty_visualization}(c), (d)). Instead, both uncertainties are more affected by the environmental noises, e.g. distant proposals depict high uncertainties. Similar results have also been observed in~\cite{feng2018leveraging}.

\begin{figure}[tpb]
	\centering
	\includegraphics[width=1\linewidth]{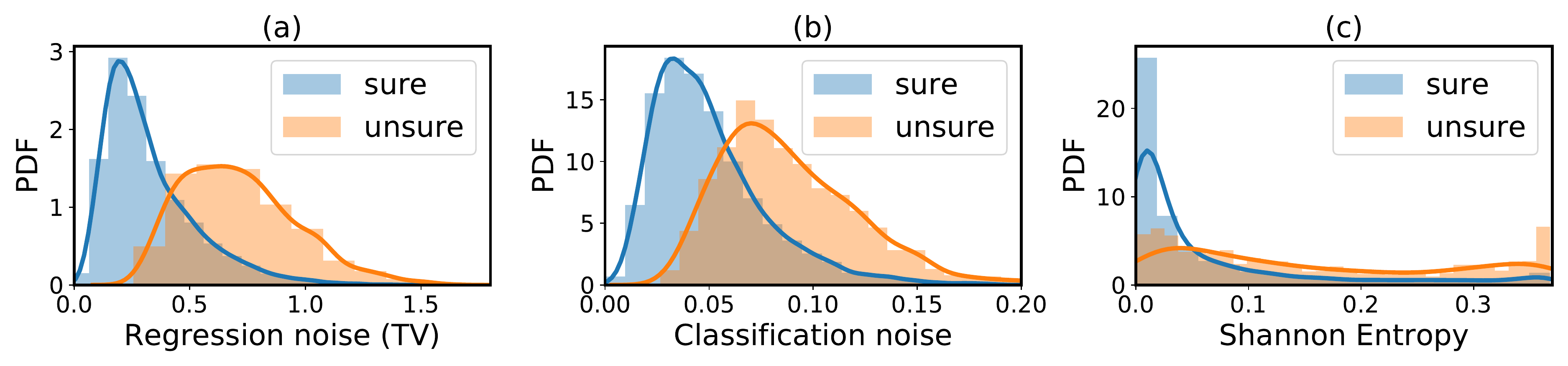}
	\caption{Histograms of the uncertainties for ``Sure" and ``Unsure" objects. The uncertainties are predicted by the LiDAR network using our Bosch dataset. (a). The total variance of the regression noise $\sigma^2_{b_z}$; (b). The classification noise for positive logit $\sigma^2_{l_z}$; (c). The Shannon entropy for the softmax scores.}\label{fig:uncertainty_sure_unsure}
\end{figure}

In order to study whether the predicted uncertainties $U_{reg}$ and $U_{cls}$  match human annotations of ``Unsure/Sure" objects, we first associate LiDAR proposals with ground truths when IoU$>0.5$, and then calculate the uncertainty histograms of associated proposals regarding on the ``Unsure/Sure" labels. As shown in Fig.~\ref{fig:uncertainty_sure_unsure}, the distributions are different between ``Unsure" objects and ``Sure" objects, with ``Unsure" detections in general showing higher $SE$, $U_{reg}$ and $U_{cls}$ scores than ``Sure" detections. To check whether the correlation in Fig.~\ref{fig:uncertainty_sure_unsure} is mainly due to effects of other environmental effects, we train linear models for $U_{reg}$ and $U_{cls}$ with independent variables ``Distance'', ``Occlusion'', ``Number of LiDAR points'' (within a bounding box), and ``Unsure'' label, where the ordinal variable ``Occlusion'' is represented by orthogonal polynomials. The resulting models have an adjusted $R^2$ of $0.43$ for $U_{reg}$ and $0.21$ for $U_{cls}$, indicating that the regression noise can be better modelled by the available parameters than the classification noise. T-tests for the regression parameter of both models result in  the p-values with $p<10^{-10}$ for all parameters. Thus, all of the parameters have a significant impact on the noise even if being corrected for the other variables. To conclude, the predictive uncertainties reflect the environmental noises which are measured by the difficulty of human annotations.

\subsection{Detection Performance}\label{subsec:detection_performance}
We study the detection performance of our proposed method in three different datasets, shown in Tab.~\ref{tab:detection_results}. Our method (``Ours") explicitly estimates both regression and classification uncertainties in the LiDAR network, and performs sampling during the training phase. It is compared with a model without the sampling mechanism (``Modelling uncertainty" in Tab.~\ref{tab:detection_results}), and the baseline model neither with sampling nor uncertainty estimation. Following~\cite{Geiger2012CVPR}, we use Average Precision to evaluate detections in the 3D space ($AP_{3D}$), in the bird's eye view ($AP_{BEV}$), as well as on the camera front-view plane ($AP_{2D}$). We group detections according to their distance to the ego-vehicle. For the Bosch and KITTI datasets, we report results up to $70$m detection distance, and set the Intersection over Union (IoU) threshold in a decreasing order, i.e. $0-30$m: IoU=0.7; $30-50$m: IoU=0.6; $50-70$m: IoU=0.5, because localizing distant objects using LiDAR data becomes very difficult due to point cloud sparsity. As for the NuScenes dataset, we set the detection distance up to $50$m, and IoU=0.5.

Tab.~\ref{tab:detection_results} shows that the networks perform similarly on the Bosch and KITTI datasets, probably due to similar sensor settings. However, all networks perform much worse on the NuScenes dataset even with smaller detection range and less strict IoU threshold, depicting the perception challenge when the number of LiDAR channels is halved. In all datasets, ``Modelling uncertainty" consistently outperforms ``Baseline" with an increase of $AP$ scores up to $7\%$. This is because probabilistic inference aids the LiDAR network to predict more accurate 3D proposals. Similar results have also been found in our previous study~\cite{feng2018leveraging}. Built upon ``Modelling uncertainty", ``Ours" further improves the detection performance in most cases (though marginal), indicating that the sampling mechanism helps networks to generalize.

\begin{table*}[tbp]
	\centering
	\resizebox{1.00\linewidth}{!}{\begin{tabular}{c|c c c|c c c|c c c}
			\Xhline{2\arrayrulewidth}
			\multirow{2}{*}{\textbf{Bosch dataset}} & \multicolumn{3}{c|}{$AP_{3D} (\%)$} & \multicolumn{3}{c|}{$AP_{BEV} (\%)$} & \multicolumn{3}{c}{$AP_{2D} (\%)$} \\ \cline{2-10}
			& $0-30$ [m] & $30-50$ [m] & $50-70$ [m] & $0-30$ [m] & $30-50$ [m] & $50-70$ [m] & $0-30$ [m] & $30-50$ [m] & $50-70$ [m] \\ \hline  
			Baseline & $88.60$ & $70.72$ & $38.36$ & $85.67$ & $69.33$ & $35.59$ & $85.32$ & $62.72$ & $27.04$ \\ 
			Modelling uncertainty & $89.11$ & $74.17$ & $42.20$ & $88.70$ & $73.18$ & $41.25$ & $88.64$ & $68.91$ & $33.40$ \\ 
			\rowcolor{lightgray!30} Modelling uncertainty + Sampling (\textbf{Ours}) & $\mathbf{89.21}$ & $\mathbf{74.40}$ & $\mathbf{42.84}$ & $\mathbf{88.92}$ & $\mathbf{73.54}$ & $\mathbf{41.75}$ & $\mathbf{88.83}$ & $\mathbf{69.46}$  & $\mathbf{34.61}$   \\ \hline	\hline
			
			\multirow{2}{*}{\textbf{KITTI dataset}} & \multicolumn{3}{c|}{$AP_{3D} (\%)$} & \multicolumn{3}{c|}{$AP_{BEV} (\%)$} & \multicolumn{3}{c}{$AP_{2D} (\%)$} \\ \cline{2-10}
			& $0-30$ [m] & $30-50$ [m] & $50-70$ [m] & $0-30$ [m] & $30-50$ [m] & $50-70$ [m] & $0-30$ [m] & $30-50$ [m] & $50-70$ [m] \\ \hline  
			Baseline & $87.24$ & $72.19$ & $37.23$ & $83.72$ & $68.43$ & $32.72$ & $84.53$ & $64.52$ & $32.32$ \\ 
			Modelling uncertainty & $87.04$ & $72.69$ & $37.64$ & $84.05$ & $69.25$ & $33.52$ & $84.64$ & $\mathbf{66.07}$ & $\mathbf{33.44}$ \\ 
			\rowcolor{lightgray!30} Modelling uncertainty + Sampling (\textbf{Ours}) & $\mathbf{87.84}$ & $\mathbf{74.15}$ & $\mathbf{39.56}$ & $\mathbf{84.79}$ & $\mathbf{70.09}$ & $\mathbf{35.01}$ & $\mathbf{87.15}$ & $65.79$ & $32.44$   \\ \hline \hline

			\multirow{2}{*}{\textbf{NuScenes dataset}} & \multicolumn{3}{c|}{$AP_{3D} (\%)$} & \multicolumn{3}{c|}{$AP_{BEV} (\%)$} & \multicolumn{3}{c}{$AP_{2D} (\%)$} \\ \cline{2-10}
			& $0-20$ [m] & $20-35$ [m] & $35-50$ [m] & $0-20$ [m] & $20-35$ [m] & $35-50$ [m] & $0-20$ [m] & $20-35$ [m] & $35-50$ [m] \\ \hline  
			Baseline & $45.01$ & $21.29$ & $13.00$ & $58.64$ & $29.99$ & $23.81$ & $55.23$ & $30.86$ & $21.77$ \\ 
			Modelling uncertainty & $47.88$ & $25.78$ & $16.21$ & $59.54$ & $33.64$ & $24.57$ & $\mathbf{57.74}$ & $34.15$ & $24.24$ \\ 
			\rowcolor{lightgray!30} Modelling uncertainty + Sampling (\textbf{Ours}) & $\mathbf{52.22}$ & $\mathbf{28.13}$ & $\mathbf{18.10}$ & $\mathbf{64.03}$ & $\mathbf{35.93}$ & $\mathbf{26.20}$ & $56.39$ & $\mathbf{35.04}$ & $\mathbf{25.90}$   \\ \Xhline{2\arrayrulewidth}
	\end{tabular}}
	\caption{Comparison of detection performance on three datasets.} \label{tab:detection_results}
	\vspace{-0.5em}
\end{table*}

\begin{table*}[tbp]
	\centering
	\resizebox{0.96\linewidth}{!}{\begin{tabular}{c c c|c c c|c c c|c c c}
			\Xhline{2\arrayrulewidth}
			\multicolumn{3}{c|}{\textbf{Test regression uncertainty}} & \multicolumn{3}{c|}{$AP_{3D} (\%)$} & \multicolumn{3}{c|}{$AP_{BEV} (\%)$} & \multicolumn{3}{c}{$AP_{2D} (\%)$} \\ \hline
			Model Nr. & Model & Sampling & Easy & Moderate & Hard & Easy & Moderate & Hard & Easy & Moderate & Hard  \\  \hline
			A & LiDAR network & No & $73.93$ & $60.23$ & $56.83$ & $\underline{90.27}$ & $77.52$ & $71.33$ & $90.79$ & $78.62$ & $72.26$ \\ 
			B & Fusion & No & $76.22$ & $61.35$ & $55.93$ & $85.90$ & $77.81$ & $71.55$ & $92.46$ & $83.69$ & $81.05$ \\
			C & Fusion & Fixed variance $\sigma_{b_z}=0.1$ ($x,y$) & $\underline{78.47}$ & $\mathbf{63.17}$ & $\mathbf{58.99}$ & $87.34$ & $\underline{78.48}$ & $\underline{73.77}$ & $92.92$ & $84.19$ & $\underline{81.56}$ \\ 
			D & Fusion & Fixed variance $\sigma_{b_z}=0.15$ ($x,y$) & $75.89$ & $61.98$ & $56.38$ & $85.18$ & $77.70$ & $71.48$ & $92.20$ & $\mathbf{85.88}$ & $81.49$ \\    
			E & Fusion & Fixed variance $\sigma_{b_z}=0.3$ ($x,y$) & $72.99$ & $59.01$ & $53.38$ & $84.78$ & $75.59$ & $70.78$ & $91.83$ & $83.29$ & $78.79$ \\    			
			\rowcolor{lightgray!30} F & Fusion & \textbf{Ours} ($x,y$) & $76.77$ & $\underline{62.04}$ & $56.41$ & $85.65$ & $77.89$ & $71.55$ & $\mathbf{94.75}$ & $83.96$ & $81.36$ \\ 
			\rowcolor{lightgray!30} G & Fusion & \textbf{Ours} ($x,y,z,w,h,l$) & $\mathbf{79.03}$ & $61.95$ & $\underline{57.29}$ & $\mathbf{90.36}$ & $\mathbf{79.58}$ & $\mathbf{74.52}$ & $\underline{93.32}$ & $\underline{84.63}$ & $\mathbf{81.78}$ \\ \hline \hline
			
			\multicolumn{3}{c|}{\textbf{Test classification uncertainty}} & \multicolumn{3}{c|}{$AP_{3D} (\%)$} & \multicolumn{3}{c|}{$AP_{BEV} (\%)$} & \multicolumn{3}{c}{$AP_{2D} (\%)$} \\ \hline
			Nr. & Model & Sampling & Easy & Moderate & Hard & Easy & Moderate & Hard & Easy & Moderate & Hard  \\  \hline
			H & LiDAR network & No & $72.51$ & $60.76$ & $\mathbf{59.79}$ & $84.98$ & $\mathbf{79.16}$ & $\mathbf{77.68}$ & $90.16$ & $81.22$ & $81.23$
			\\ 
			I & Fusion & No & $75.91$ & $63.35$ & $58.07$ & $85.44$ & $77.90$ & $73.33$ & $92.32$ & $84.08$ & $81.51$ \\  
			J & Fusion & Fixed variance $\sigma_{l_z}=0.17$ & $\underline{78.96}$ & $\underline{63.86}$ & $58.31$ & $\underline{86.71}$ & $78.65$ & $73.97$ & $\mathbf{94.88}$ & $\underline{86.26}$ & $\mathbf{83.54}$ \\ 
			\rowcolor{lightgray!30} K & Fusion & \textbf{Ours} & $\mathbf{79.71}$ & $\mathbf{65.24}$ & $\underline{59.55}$ & $\mathbf{88.49}$ & $\underline{78.99}$ & $\underline{74.18}$ & $\underline{93.05}$ & $\mathbf{86.46}$ & $\underline{82.03}$ \\ \Xhline{2\arrayrulewidth}
	\end{tabular}}
	\caption{Ablation study on the KITTI dataset (Best performance is marked in bold, second best with underline).} \label{tab:ablation_study}
	\vspace{-1.0em}
\end{table*}

\noindent \textbf{Ablation Study:} We additionally conduct ablation study to validate the proposed method. Tab.~\ref{tab:ablation_study} shows the detection performance on the KITTI dataset. We divide data into Easy, Moderate, and Hard settings~\cite{Geiger2012CVPR}, and use IoU=0.7 for all settings. Models A, H are LiDAR networks which only model regression uncertainty and classification uncertainty, respectively. The fusion networks B-G are built upon Model A, while Models I-K upon Model H. To verify the advantage of sampling from the learned uncertainties, we additionally train fusion networks by sampling with fixed standard deviation for the regression uncertainty (Models C-E) and the classification uncertainty (Model J). The fixed deviation for regression uncertainty $\sigma_{b_z}$ is increased from $0.1$m to $0.3$m, and the deviation for classification uncertainty $\sigma_{l_z}$ is selected as the mean value of the learned uncertainty. 

From the table we have three observations. First, our proposed sampling mechanism consistently improves the fusion performance either for regression (comparing Models F, G with Model B) or classification (comparing Model K with I), and sampling more regression variables could bring better detection accuracy (comparing G with F). Second, sampling with the fixed small deviation (Model C) improves the detection accuracy due to the data augmentation advantage. However, increasing the fixed deviation may harm the performance (Models D and E). In fact, sampling at $\sigma_{b_z}=0.3$m (Model E) even underperforms the LiDAR-only detector (Model A). Similar results have also been found in\mbox{~\cite{haase2019estimating}}. Therefore, the fixed-sampling approach is an ad-hoc solution and requires tedious hyper-parameter tuning. In contrast, our method of sampling learned uncertainties (Models F, G and K) avoids such process. It generates diverse training data which corresponds to complex environmental noises (as shown in \mbox{Sec.~\ref{subsec:uncerstanding_uncertainty}}), and provides competitive or better detection performance, making it more robust than the fixed sampling method. Finally, fusing LiDAR data with RGB camera information largely improves 2D detection results (e.g. comparing Model A and B on $AP_{2D}$), but may degrade the BEV performance from the LiDAR-only network (e.g. Models E and F on $AP_{BEV}$) due to the fusion network design: the network implicitly learns sensor alignment between LiDAR top-down view and camera front-view, which makes training the 3D bounding box estimation challenging. One remedy for this problem is to project camera images onto the LiDAR top-down view before fusion, similar to~\cite{liang2018deep}. 

\subsection{Robustness Testing} \label{subsec:robustness_testing}
So far, we have shown how the proposed method works in datasets with well-aligned sensor settings. When deploying a multi-modal object detector online, however, sensor mismatch could occur due to different timestamp (i.e. temporal misalignment) or calibration errors (i.e. spatial misalignment), which may drastically degrade the perception performance. An ideal object detector should not only perform well in good conditions, but show its robustness against sensor misalignment as well. 

\begin{figure}[tpb]
	\centering
	\begin{minipage}{0.42\linewidth}
		\centering
		\includegraphics[width=1.0\linewidth]{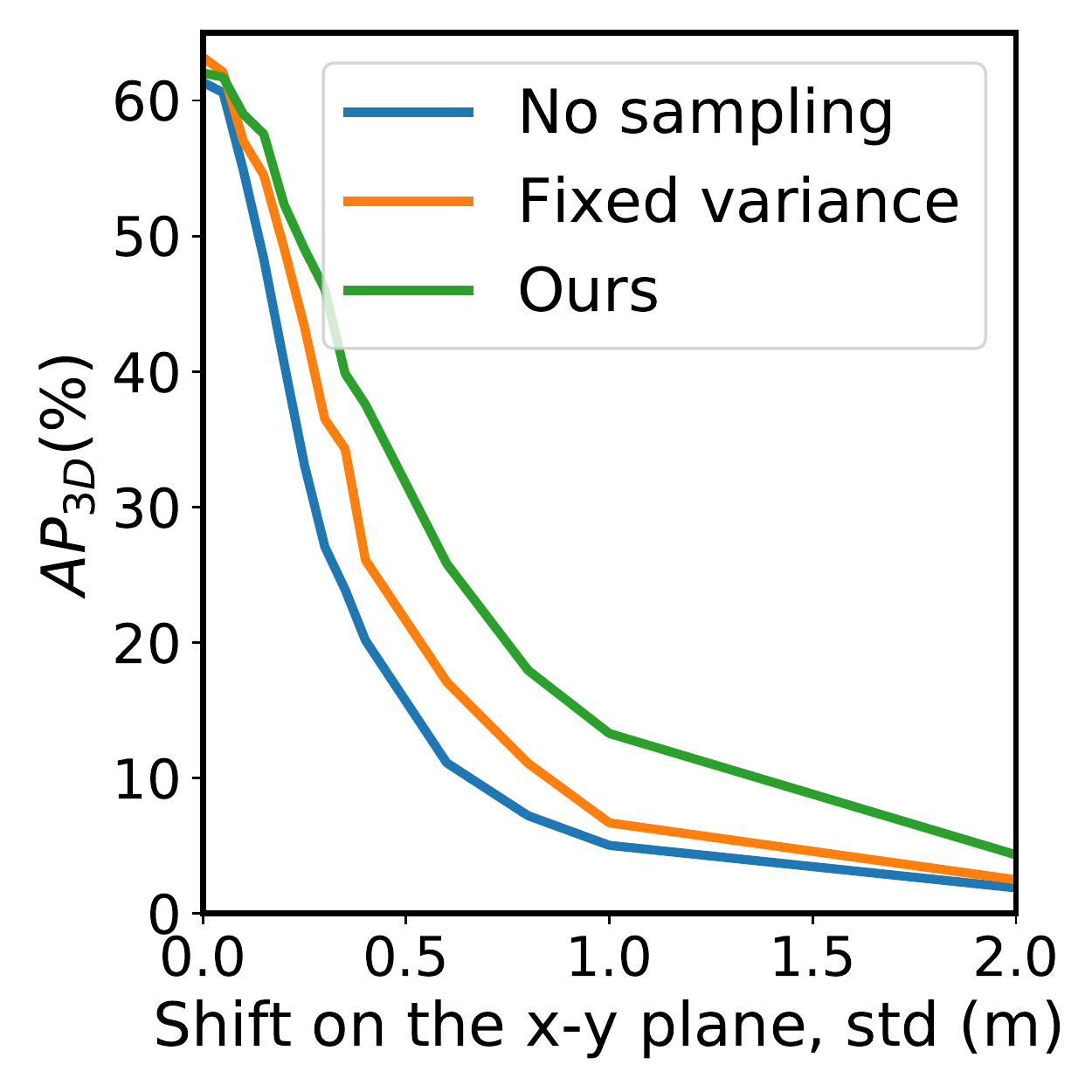}
	\end{minipage}
	\hskip 0.5pt
	\begin{minipage}{0.5\linewidth}
		\centering
		\caption{\small Robustness testing. We randomly shift the LiDAR point clouds in the horizontal plane following the Gaussian distribution to simulate the sensor temporal misalignment. The horizontal axis represents the standard deviations, and the vertical axis represents the detection performance on the KITTI dataset.}\label{fig:robustness_testing}
	\end{minipage}
	\vspace{-1.5em}
\end{figure}

In this work, we evaluate the method's robustness against the temporal misalignment. we follow~\cite{shin2019roarnet} and simulate the misalignment by randomly shifting all LiDAR point clouds in a frame following Gaussian distribution with zero means and increasing deviations, while keeping cameras as reference. We compare among three models: the fusion network trained without sampling, the one with fixed sampling ($\sigma_{l_z}=0.1$m), and the one with our sampling mechanism. All models are trained with the clean KITTI \textit{train} set, and tested with the misaligned \textit{val} set. Fig.~\ref{fig:robustness_testing} reports the 3D detection performance in the KITTI ``Moderate" setting. At the same shifting level, our method largely outperforms the models without sampling up to nearly $20\%$ $AP$ or the one with fixed sampling, showing its high robustness against noisy data. Though we only conduct experiments with the KITTI dataset, similar results are expected in other datasets as well (such as Bosch and NuScenes).

\section{Conclusion and Discussion}
\label{sec:conclusion}
We have presented our probabilistic two-stage multi-modal object detection network that fuses LiDARs and RGB cameras. The method proposes to predict 3D proposals from the LiDAR branch, and to combine the regional LiDAR and camera features with a light-weight fusion network. We explicitly model classification and regression uncertainties in the LiDAR network, and leverage those uncertainties to train the fusion network. We evaluate our method on three datasets with real-world driving scenarios. Experimental results show that the predicted uncertainties reflect complex environmental uncertainties reflected by the difficulty of human annotators to label certain objects. Furthermore, modelling uncertainties helps to improve the detection accuracy and the network's robustness, especially in the sensor misalignment situation. 

In this work, we only model uncertainties in the LiDAR branch. It is a very interesting future work to model uncertainties in the image backbone and the fusion network. In addition, to reduce the computational cost of our fusion network for online deployment, we will introduce the quantization technique~\cite{enderich2019learning} into our method.

\section*{Acknowledgment}
We thank our colleague William H. Beluch for the suggestions and fruitful discussions.

\bibliographystyle{IEEEtran}
\bibliography{bibliography}

\end{document}